\documentclass[10pt,twocolumn,letterpaper]{article}

\usepackage{wacv}
\usepackage{times}
\usepackage{epsfig}
\usepackage{graphicx}
\usepackage{amsmath}
\usepackage{amssymb}
\usepackage{booktabs}

\def\wacvPaperID{1856} 

\wacvalgorithmstrack   

\wacvfinalcopy 

\ifwacvfinal
\usepackage[breaklinks=true,bookmarks=false]{hyperref}
\else
\usepackage[pagebackref=true,breaklinks=true,colorlinks,bookmarks=false]{hyperref}
\fi

\begin{document}

\title{Segmentation-free Direct Iris Localization Networks}

\author{Takahiro Toizumi$^1$, Koichi Takahashi$^1$ and Masato Tsukada$^{1,2}$ \\
$^1$NEC corporation, $^2$University of Tsukuba \\
{\tt\small t-toizumi\_ct@nec.com, koichi.takahashi@nec.com, tsukada@iit.tsukuba.ac.jp}
}

\maketitle
\thispagestyle{empty}

\begin{abstract}
This paper proposes an efficient iris localization method without using iris segmentation and circle fitting. Conventional iris localization methods first extract iris regions by using semantic segmentation methods such as U-Net. Afterward, the inner and outer iris circles are localized using the traditional circle fitting algorithm. However, this approach requires high-resolution encoder-decoder networks for iris segmentation, so it causes computational costs to be high. In addition, traditional circle fitting tends to be sensitive to noise in input images and fitting parameters, causing the iris recognition performance to be poor. To solve these problems, we propose an iris localization network (ILN), that can directly localize pupil and iris circles with eyelid points from a low-resolution iris image. We also introduce a pupil refinement network (PRN) to improve the accuracy of pupil localization. Experimental results show that the combination of ILN and PRN works in 34.5 ms for one iris image on a CPU, and its localization performance outperforms conventional iris segmentation methods. In addition, generalized evaluation results show that the proposed method has higher robustness for datasets in different domain than other segmentation methods. Furthermore, we also confirm that the proposed ILN and PRN improve the iris recognition accuracy.
\end{abstract}

\section{Introduction}
Iris recognition is one of the most accurate and reliable technologies for biometrics. It is applied to various fields including forensic science, border control, biometric payment systems, and biometric unlock on cell phones. To promote such applications, iris recognition systems require not only environmental robustness but also high-speed operation.

Although a lot of image processing tasks have been replaced with deep learning methods, iris recognition systems have kept their traditional frameworks. The process of iris recognition can be divided into imaging, segmentation, localization, normalization, feature extraction, and matching. Iris segmentation extracts a map of the iris area from a single eye input image. Afterward, iris localization calculates the center and radius of the pupil and iris from the iris segmentation map. The extracted pupil and iris circles are used for the normalization and feature extraction. Iris segmentation in particular is easily affected by the iris imaging environment, such as the lighting conditions. Thus, many CNN-based semantic segmentation methods robust to environmental variations or non-cooperative situations have been proposed \cite{jalilian2017iris, lakra2018segdensenet, lian2018attention, lozej2018end, wang2020towards}. In an iris localization challenge held at IJCB 2021, all methods used segmentation models \cite{wang2021nir}. Iris segmentation and localization methods tend to be focused on improving robustness.

\begin{figure}[t]
\begin{center}
   \includegraphics[width=0.92\linewidth]{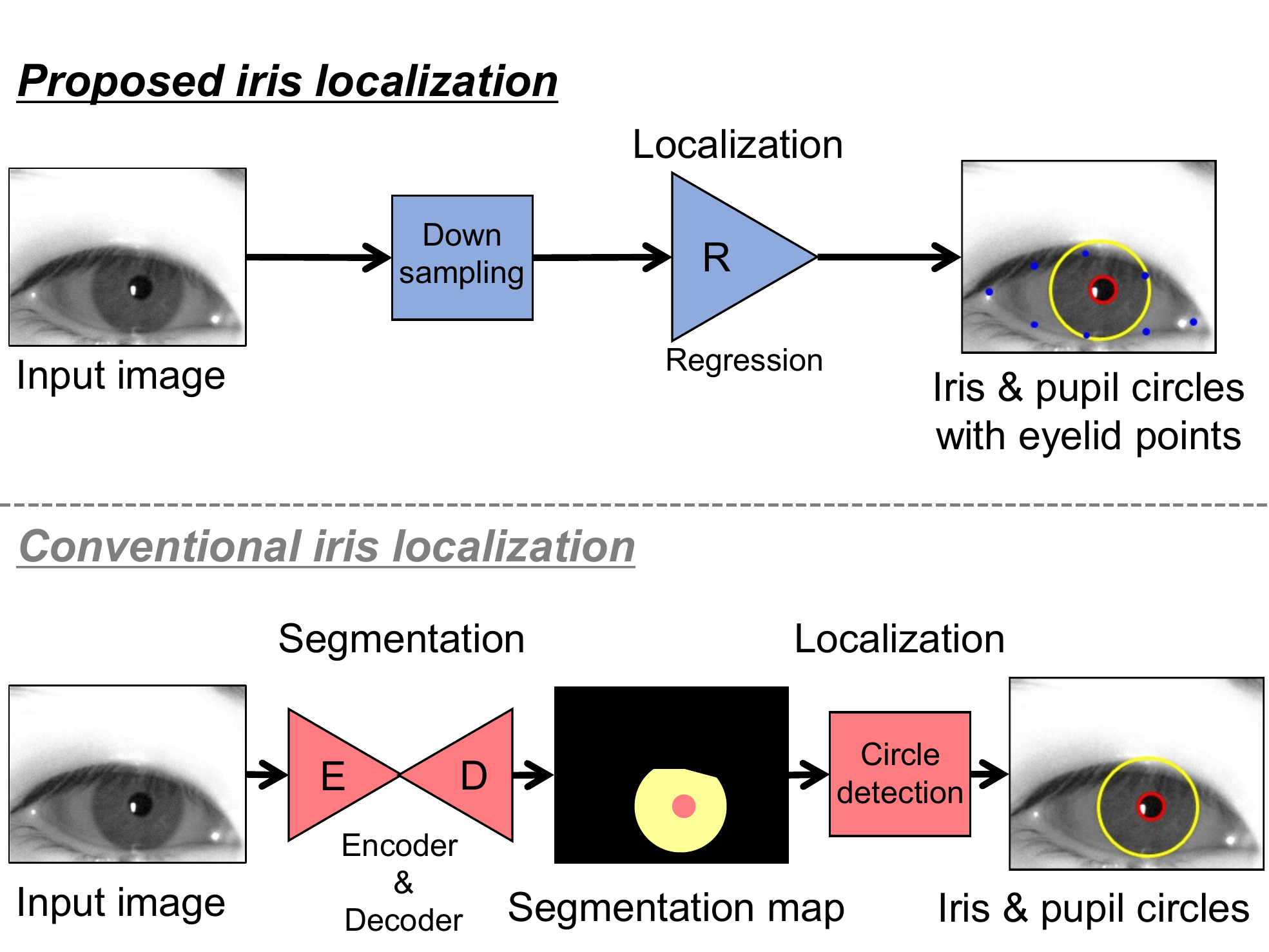}
\end{center}
   \caption{Difference between proposed and conventional iris localization methods. Proposed method directly extracts pupil and iris circles with eight eyelid points using regression network without iris segmentation. It localizes iris regions within 34.5 ms computational time on CPU while maintaining higher detection performance than conventional method.}
\label{fig:top}
\end{figure}

Iris localization methods based on iris segmentation have room for improvement in three areas: processing speed, annotation, and robustness. In terms of processing speed and annotation, semantic segmentation methods generally have higher computational complexity and higher annotation cost than other tasks such as regression or classification. The computational complexity increases noticeably when extracting high-resolution iris segmentation maps, usually with a size of 640 $\times$ 480, without a GPU. The processing time is one of the most important issue on iris on the move (IOTM) \cite{matey2006iris}. IOTM requires faster processing so that subjects can keep walking in the recognition gate. The annotation cost also increases when introducing new dataset or re-designed iris segmentation maps. In terms of robustness, post-processing of iris segmentation results may also contribute to limiting the robustness of the overall iris localization. Although iris segmentation focuses on improving robustness, iris localization still relies on the traditional binarization and circle fitting algorithm. This traditional algorithm can cause the iris localization and recognition performance to be poor, especially when iris segmentation maps are ambiguous because of noisy images. In addition, the performance of the traditional iris localization is sensitive to its parameters. 

To achieve higher speed, robustness and annotation efficiency, we newly designs the iris landmarks for iris localization and recognition. Unlike general facial landmarks, our proposed iris landmarks include both circles and points to skip the circle fitting processing. In addition to the landmark design, we propose an iris localization network (ILN) to directly detect pupil and iris outer circles using a deep regression network to achieve segmentation-free iris localization. ILN directly outputs pupil and iris circles and eyelid points from a down-sampled low-resolution image as shown in Figure \ref{fig:top}. It localizes iris circles without high-resolution iris segmentation maps, so the iris localization is even faster than the conventional segmentation-based methods. In addition, we apply a pupil refinement network (PRN) to improve the accuracy of pupil circle localization using cropped iris images. Our configuration can largely reduce the annotation cost because it needs only two circles and eight points as a ground truth. Eyelid, eyelash, and specular masks are generated by using the eight points and pixel values for the iris region instead of the segmentation results. Experimental results show that the combination of ILN and PRN works faster and more accurately than the conventional iris segmentation methods. Furthermore, we also confirm that the proposed method can improve the iris recognition accuracy compared with the conventional methods.

\section{Related work}

Several iris recognition methods have been proposed before and after the CNN development \cite{al2018, cho2017periocular, gangwar2016, guo2019, hernandez2018, hu2020, lee2019deep, nguyen2018, rodriguez2011segmentation,ross2012matching, yang2021, zhao2017, zhao2018}. In these methods, the iris image is normalized by Daugman's rubber sheet model \cite{daugman2004} using localization results. Iris features are extracted from the normalized images \cite{al2018,  gangwar2016, nguyen2018, yang2021, zhao2017}. Some studies have extracted recognizable features from detected iris or periocular bounding boxes instead of the segmentation and normalized images \cite{cho2017periocular, guo2019, hernandez2018, hu2020, lee2019deep, proencca2019segmentation, rodriguez2011segmentation, zhao2018}. In general, the normalized image leads to better performance than the bounding boxes because Daugman's model is independent of pupil size. The iris feature extractor also uses non-iris region masks generated by segmentation maps. Lozej \etal. \cite{lozej2019influence} investigated the influence of normalization and eyelid region masking, and they confirmed that these processes improve the iris recognition performance.

Iris recognition requires determining iris regions and its coordinates by iris segmentation and localization. For iris segmentation, several methods have been proposed including feature-based methods \cite{othman2016osiris, tan2010efficient, zhao2015accurate} and CNN-based methods \cite{arsalan2018irisdensenet, fang2020open, hofbauer2019exploiting, jalilian2017iris, lakra2018segdensenet, lian2018attention,  lozej2018end, wang2020towards, zhao2021detection}. While feature-based methods are fast, their performance is reduced by environmental variations such as glasses reflections or off-axis eyes. CNN-based methods are robust to environmental variations, but they tend to be slower due to their architectures such as U-Net \cite{ronneberger2015u}. Some methods attempt to reduce the computational cost by using light weight CNN \cite{fang2020open} or small input images \cite{zhao2021detection}. However, the localization performance of these methods are yet limited by the post processing of localization. 

CNN-based object detection is a faster task than segmentation \cite{liu2016ssd, proencca2019segmentation, redmon2016you, ren2016faster}. The object detection task extracts bounding boxes of target objects from an input image. Li \etal. \cite{li2019efficient} proposed an efficient iris localization using a light weight Faster RCNN \cite{ren2016faster} to improve the computational efficiency. Since their process can be split into bounding box detection and circle localization, further simplification can be expected. Instead of bounding box detection, CircleNet \cite{yang2020circlenet} detects bounding circles in an object detection framework. It estimates a likelihood map to detect coarse positions of target objects, and it finely detects the centers and radii of bounding circles using regression.

In the research field of face recognition, direct or few-level regression using CNNs has been proposed for facial landmark detection \cite{Feng_2018_CVPR, sun2013deep, Zhang2016joint, zhou2013extensive}. These methods directly detect facial landmarks under the assumption that the landmarks are always included in input images. This assumption allows facial landmark detection to skip generating high-resolution segmentation maps. In addition, facial landmark detection methods use refinement processes to improve their performance, and it achieves few pixel-level accuracies.

\begin{figure*}
\begin{center}
\includegraphics[width=0.95\linewidth]{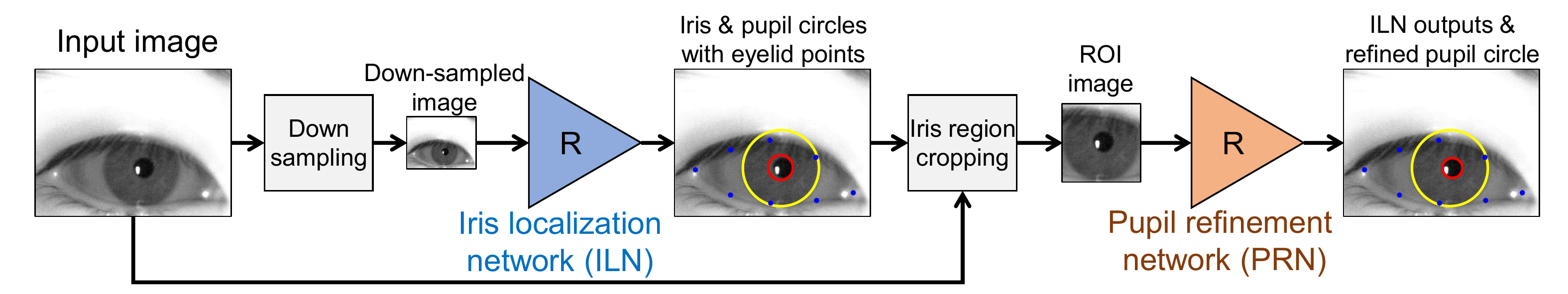}
\end{center}
   \caption{Our proposed iris localization method. Method has iris localization network (ILN) and pupil refinement network (PRN).}
\label{fig:method}
\end{figure*}

In this paper, the proposed ILN creates no iris segmentation maps, and directly localizes the pupil and iris circles based on the landmark detection scheme under the assumption that the input iris image always contains one iris. Because this paper is the first attempt to apply landmark detection for iris recognition, we select pupil and iris circles as the target. It can avoid traditional noise sensitive circle fitting. In addition to the circles, ILN detects eyelid points simultaneously. The eyelid points are used to create eyelid masks for iris recognition. Moreover, the accuracy in localizing pupil circles is sensitive to the iris recognition performance because misalignment between the ILN output and ground truth changes normalized rubber sheet images drastically. Thus, we introduce PRN for the fine localization of pupil circles. Although we mainly select circles and points in this paper, our method can easily be extended to ellipse \cite{ryan2008adapting} or any shape \cite{daugman2007new, shah2009iris} localization. 

The contributions of this paper are as follows.
\begin{itemize}
 \setlength{\itemsep}{0cm}
 \item We newly design iris landmarks for iris recognition. The proposed landmarks are two circles and eight eyelid points, which are different from points or circle only detectors \cite{sun2013deep, yang2020circlenet}. This configuration achieves accurate iris localization with lower annotation cost.
 \item The proposed iris localization method is fast and robust. The proposed ILN localizes circles directly from down-sampled images without circle fitting. The proposed PRN further improves the pupil localization accuracy, which is important for iris recognition.
\end{itemize}

\section{Iris localization and pupil refinement}
We explain the proposed iris localization networks in this section. The proposed method utilizes down-sampled input, a regression model, the region of interest (ROI) and refinement regression to achieve fast and accurate localization.

Figure \ref{fig:method} shows our proposed iris localization method. The method uses an iris localization network (ILN) and pupil refinement network (PRN). To avoid noise and parameter sensitive circle fitting processing, ILN directly localizes circles and eyelid points from a down-sampled input image. In addition, to boost iris recognition performance, PRN re-localizes a pupil circle with few-pixel accuracy from a ROI image. The ROI image is cropped from an original image using the iris circle localized by ILN. Note that the ROI image is only used for the pupil refinement, and the iris rubber sheet is extracted from the original input image with localized circles for iris recognition. Thus, the ROI is not directly affected to the iris rubber sheet extraction.

\begin{figure}[t]
\begin{center}
   \includegraphics[width=0.95\linewidth]{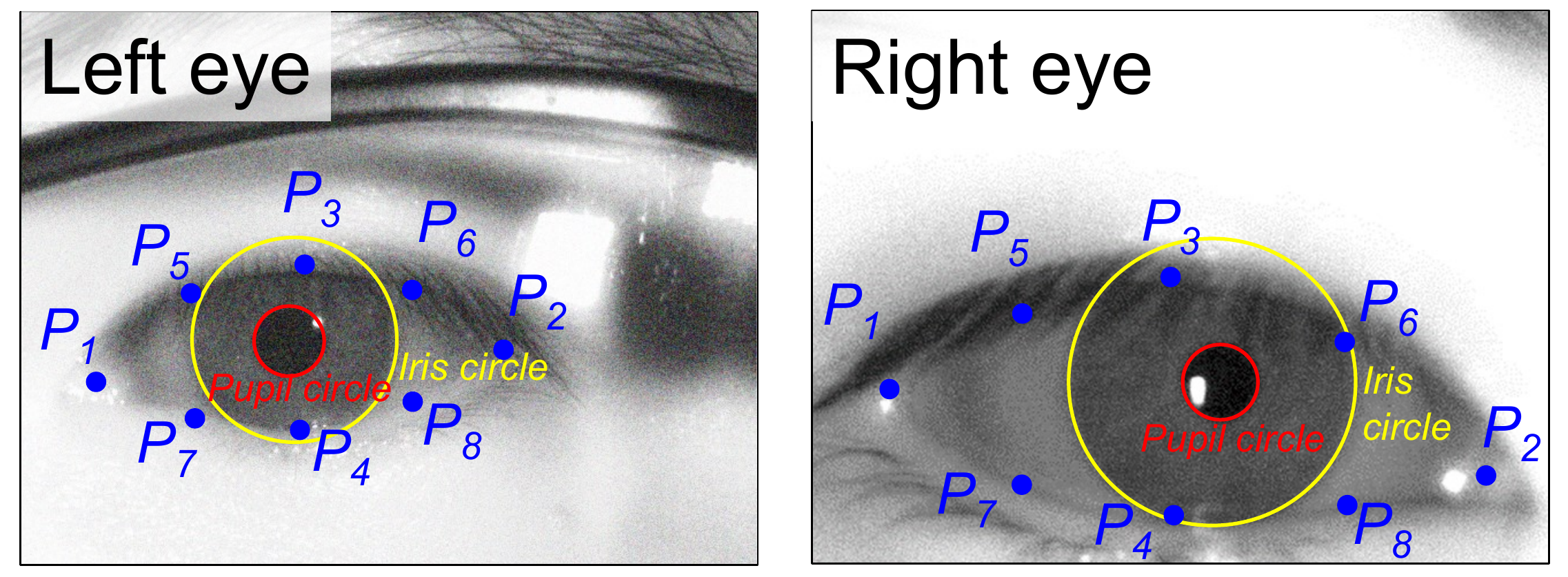}
\end{center}
   \caption{Proposed location for circles and points in left and right eye images. Position of $P_1$ is on left side in images for both eyes.}
\label{fig:eyes}
\end{figure}

Figure \ref{fig:eyes} shows configurations of target circles and points. The proposed method is designed to treat the circles and points as having the same location configurations in left and right iris images. The position of $P_1$ is the left side corner in both images for each eye. It does not refer to the inner or outer eyelid corner. These circles and points are summarized to a target vector $k$. The circle is represented as a vector with $x$ and $y$ coordinates and the radius $r$. A point is represented as a vector with $x$ and $y$ coordinates. Thus, the target vector $k$ has 22 dimensions including 6 dimensions for the pupil and iris circles and 16 dimensions for the 8 eyelid points. 

The architectures of ILN and PRN are constructed on the basis of the VGG network structure \cite{simonyan2014very}. For ILN, we define the input image scale and the channel width by $s$ and $m$, respectively. The down-sampled image size is defined as 640$s$ $\times$ 480$s$ when the original size is 640 $\times$ 480. The network width is reduced in the channel direction by multiplying the channel width $m$. The last layer is replaced by a linear layer with output vector length $d$. The output vector has a length of $d$ = 22 for ILN. 

PRN uses an ROI image as an input. The ROI image is cropped from the original input image using the detected iris center and radius in the ILN output. The ROI image is resized to a size of $128 \times 128$ pixels and fed into the network. PRN has no resize parameter, and the network width $m$ is shared with ILN. The output vector of PRN has a length of $d$ = 3, which includes the pupil circle elements.

Unlike iris segmentation models, our regression model can emphasize the importance of output elements by using a weighted loss function. Weights are given for each element in the output vector. In ILN, the pupil and iris elements are given higher weights than the eyelid elements because the accuracy of iris localization is important for iris recognition. PRN uses equal weights for all three elements of the output vector $f_{PRN}(I)$. The loss functions $\mathcal{L}_{ILN}$ and $\mathcal{L}_{PRN}$ of the two proposed networks are written as:
\begin{equation}
    \mathcal{L}_{ILN}= \sum^{d=22}_i{w_i|| f_{ILN}(I)_i - k_i||_1}, \label{eq:loss1}  
\end{equation}
\begin{equation}
   \mathcal{L}_{PRN} = \sum^{d=3}_i{w_i||f_{PRN}(I)_i - k_i||_1},  \label{eq:loss2}
\end{equation}
where $I$ is the input image. $f(I)_i$, $k_i$, and $w_i$ denote the $i$-th element of the output vector, target ground truth vector and weight, respectively. $d$ is the length of the output vector $f(I)$. We use L1-norm for the loss function.

\begin{figure}[t]
\begin{center}
   \includegraphics[width=0.99\linewidth]{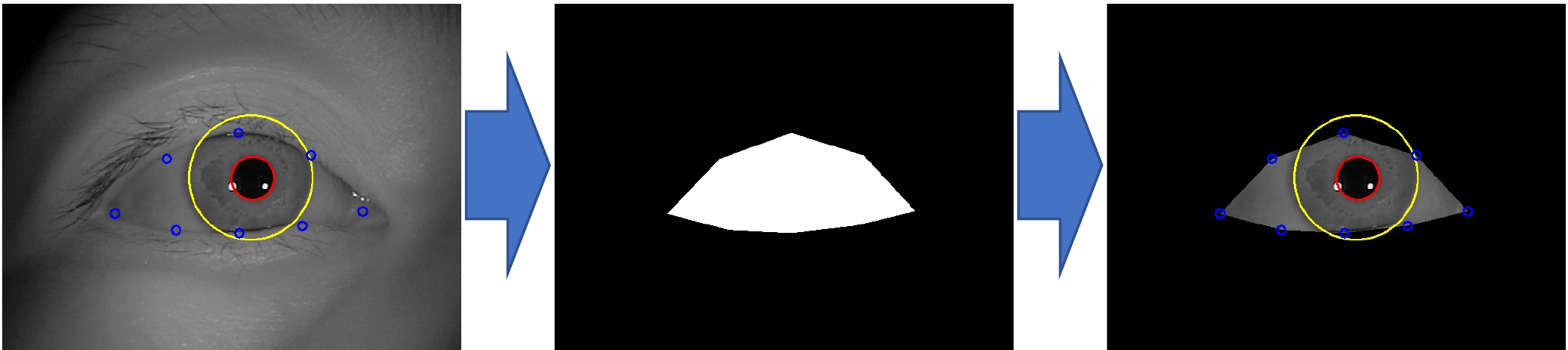}
\end{center}
   \caption{Mask generation for iris recognition. Iris and eyelid masks are generated by localized circles and points.}
\label{fig:mask}
\end{figure}

\begin{figure}[t]
\begin{center}
   \includegraphics[width=0.90\linewidth]{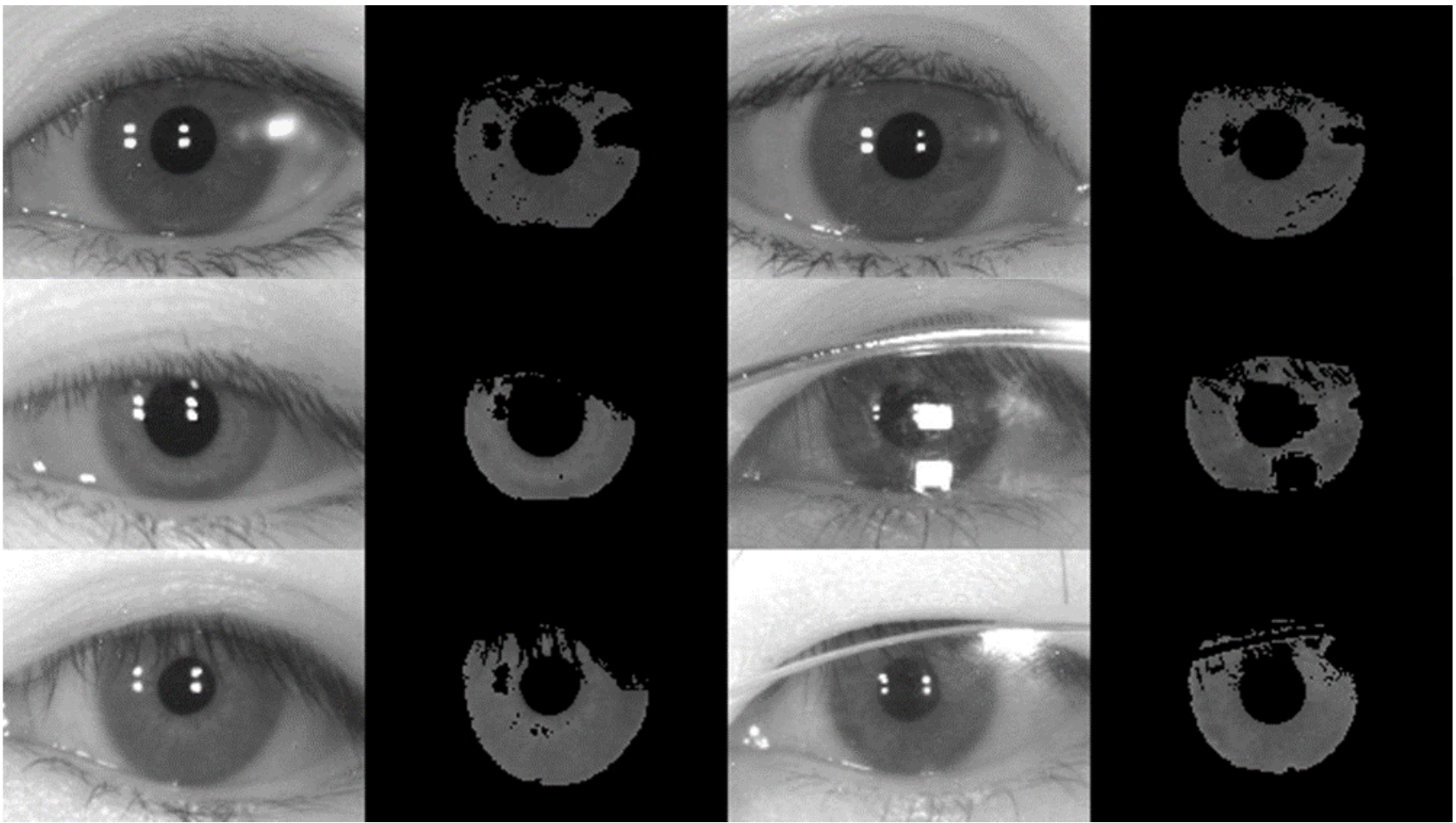}
\end{center}
   \caption{Eyelid, eyelash and specular mask results with proposed method and simple anomaly detection.}
\label{fig:seg}
\end{figure}

For iris recognition, it is necessary to exclude eyelid regions from the detected iris circles. Our method creates an eyelid mask from the eight detected eyelid points. As shown in Figure \ref{fig:mask}, the eyelid mask is created by connecting the eight localized eyelid points. The reason for the eight points is to maintain the trade-off between having a sufficient number of points for creating the mask and a minimum number of points for reducing the annotation cost. Note that our method does not directly remove the specular region on the iris surface or eyelash occlusion. Thus, we adapt simple anomaly detection by interquartile for pixel values in the localized iris region to mask the specular and eyelash regions. The results of masking are shown in Figure \ref{fig:seg}. The specular and eyelash regions can be masked by this simple anomaly detection without segmentation and its annotation.

\section{Experiments}
We evaluated the proposed method in five types of experiments. In an ablation study, we evaluated the parameter dependency of the proposed method. Afterward, dataset dependency and generalization performance were evaluated. We also show the extension to ellipse localization. Finally, we demonstrate the effectiveness of the proposed method for iris recognition using a feature extractor.

Table \ref{tab:dataset} shows details on the datasets used for the evaluations. The datasets ware CASIA v4-thousand, CASIA v4-distance, IITD, MMU1, MMU2 and CASIA v4-twins \cite{mmudataset, casia, kumar2010comparison}. For CASIA v4-distance, we extracted eye regions from the dataset images using manually annotated eye points. The dataset included only NIR images, no RGB images. This is because we assume that the proposed iris localization model will be used for iris recognition in the NIR band. The images in all datasets were resized to 640 $\times$ 480 with the aspect ratio kept the same before being fed into the proposed model. For images without a 4:3 aspect ratio, we replicated the lower aspect borders and aligned the aspect ratio to 4:3 before resizing. We split CASIA v4-thousand, CASIA v4-distance and IITD datasets into about 80\% for training and 20\% for test data. The number of subjects and images are shown in Table \ref{tab:dataset}. For MMU1 and MMU2, we used all datasets for testing in a generalization evaluation. For only CASIA v4-twins, we used only the last 20\% for testing this evaluation. We manually annotated the ground truth location for all datasets. Note that to guarantee the accuracy of our annotation, we additionally evaluated the recognition performance in section \ref{sec:recognition}, and confirmed that it was approximately consistent to localization results.

For evaluation, we selected five conventional methods and compared their performance with that of the proposed method. We used IrisParseNet \cite{wang2020towards}, OSIRIS \cite{othman2016osiris}, CC-Net \cite{fang2020open}, IrisDenseNet \cite{arsalan2018irisdensenet} and BiSeU-Net \cite{zhao2021detection}. IrisParseNet was selected as the state-of-the-art method for iris localization. OSIRIS was selected as the most commonly used non-training method. Because OSIRIS works without a training dataset, its performance has less dependency on the dataset domain than training-based methods. CC-Net was selected as the most efficient iris segmentation method. IrisDenseNet and BiSeU-Net were selected as the state-of-the-art methods for iris segmentation. The segmentation performances of these methods outperform IrisParseNet on some datasets. However, the localization performance was not evaluated for these methods. Thus, we evaluated these methods on localization metrics.

ILN and PRN were trained independently. For training ILN, each input image $I$ was normalized to a range of [0.0, 1.0] and resized with a scale parameter $s$. The ground truth vector $k$ was normalized on the basis of the mean and standard deviation as $k'$ = $(k-\mu)/\sigma$. The mean values $\mu$ were (320, 240) for each ($x$, $y$) coordinate including circle centers and eyelid points, and (50, 120) for the pupil and iris radius. The standard deviation values $\sigma$ were set to 1/6 of the mean values. Note that $k$ is invariant to the scale parameter $s$. We trained the model with a batch size of 128 and 100,000 iterations. We used stochastic gradient descent (SGD) as the optimizer with a learning rate of 0.001 and weight decay of 0.00001. The learning rate was switched to 0.0001 after 90,000 iterations. The loss weights were set to 3.0 for the six elements of the pupil and iris circles and 1.0 for all others. For training PRN, each ROI image was cropped with 1.2 times the size of the iris outer circle using the ground truth. The ROI images were resized to 128 $\times$ 128 pixels, and the pixel values were normalized to the [0.0, 1.0] range. The ground truth vector was transformed into a vector on the ROI image coordinates and normalized using the means (64, 64, 20) and standard deviations (10, 10, 10) of the ROI image coordinates. The loss weights were set to 1.0 for the $x$, $y$, and $r$ of the output pupil circle. All other parameters were the same as ILN. In the test phase, a predicted vector was inversely transformed to the original coordinates and the pupil elements were replaced with the transformed vector.

\begin{table}
\caption{Evaluation datasets and train-test split.}
\label{tab:dataset}
\begin{center}
\small
\begin{tabular}{lcc}
\hline 
Dataset & Subjects & Images \\
 & train / test & train / test \\
\hline 
CASIA v4-thousand \cite{casia} & 800 / 200 & 16,000 / 4,000 \\
CASIA v4-distance \cite{casia} & 110 / 32 & 4,034 / 1,100 \\
IITD \cite{kumar2010comparison} & 180 / 44 & 1,800 / 440 \\
MMU1 \cite{mmudataset} & -- / 45 & -- / 450 \\
MMU2 \cite{mmudataset} & -- / 100 & -- / 995 \\
CASIA v4-twins \cite{casia} & -- / 40 & -- / 688 \\
\hline 
\end{tabular}
\end{center}
\end{table}

We applied data augmentation in the training using Gaussian blur, brightness-contrast, vertical and horizontal shifts, scaling, and rotation with a probability of 50\% for each. For Gaussian blur, the value of $\sigma$ was applied in a uniform distribution in the range of [1.0, 25.0] pixels for the original sizes of the images. The brightness and contrast were applied in uniform distributions of $\pm$ 20\% for each. Vertical and horizontal shifts were applied uniformly in a range that included the iris circle in the images. Scaling was applied in the range of [0.3, 2.0]. Rotation was applied in the range of $\pm$ 20 degrees. If an image contained out-of-frame areas due to shifting or scaling, the border pixels were replicated. In the training of PRN, we applied the same parameters for data augmentation without shifting. To shift a cropped image, the iris center was sampled by using the standard deviation of ILN's results from the ground truth.

To evaluate the iris localization, we used the normalized Hausdorff distance \cite{wang2020towards} between the estimated circle $C$ and the ground truth circle $G$ as an evaluation metric. The Hausdorff distance is defined as: 
\begin{equation}
\begin{split}
 H(G,C) = {\rm max} \{ \sup_{g \in G} \inf_{c\in C} ||g-c||, \sup_{c\in C} \inf_{g\in G} ||c-g|| \}, 
 \end{split}
\end{equation}
where $c$ and $g$ are arbitrary points on the predicted circle $C$ and the ground truth circle $G$, respectively. In addition, the Hausdorff distance was normalized by the ground truth eye width (distance between $P_1$ and $P_2$ in Figure \ref{fig:eyes}) to remove the influence of the different scales and iris sizes. A smaller normalized Hausdorff distance means a higher shape similarity and a higher localization accuracy.

\subsection{Ablation study}

We evaluated the effect of the hyperparameters and the pupil refinement defined in the proposed method. We used CASIA v4-thousand for an ablation study. The two hyperparameters $s$ and $m$ were selected from (0.1, 0.2, 0.5) for $s$ and (1.0, 0.5, 0.25, 0.125) for $m$. The combinations of $s$ and $m$ were selected so that the calculation time was less than 150 ms on a CPU (Intel Xeon CPU E3-1280 v5) with a single thread. To evaluate the pupil refinement (ILN + PRN), we used ($s$, $m$) = (0.2, 0.25) for the ILN model. We also generated an ensemble model using two models trained with the same parameters ($s$, $m$) = (0.2, 0.25) and different initialization for comparison. 

\begin{figure}[t]
\begin{center}
   \includegraphics[width=0.90\linewidth]{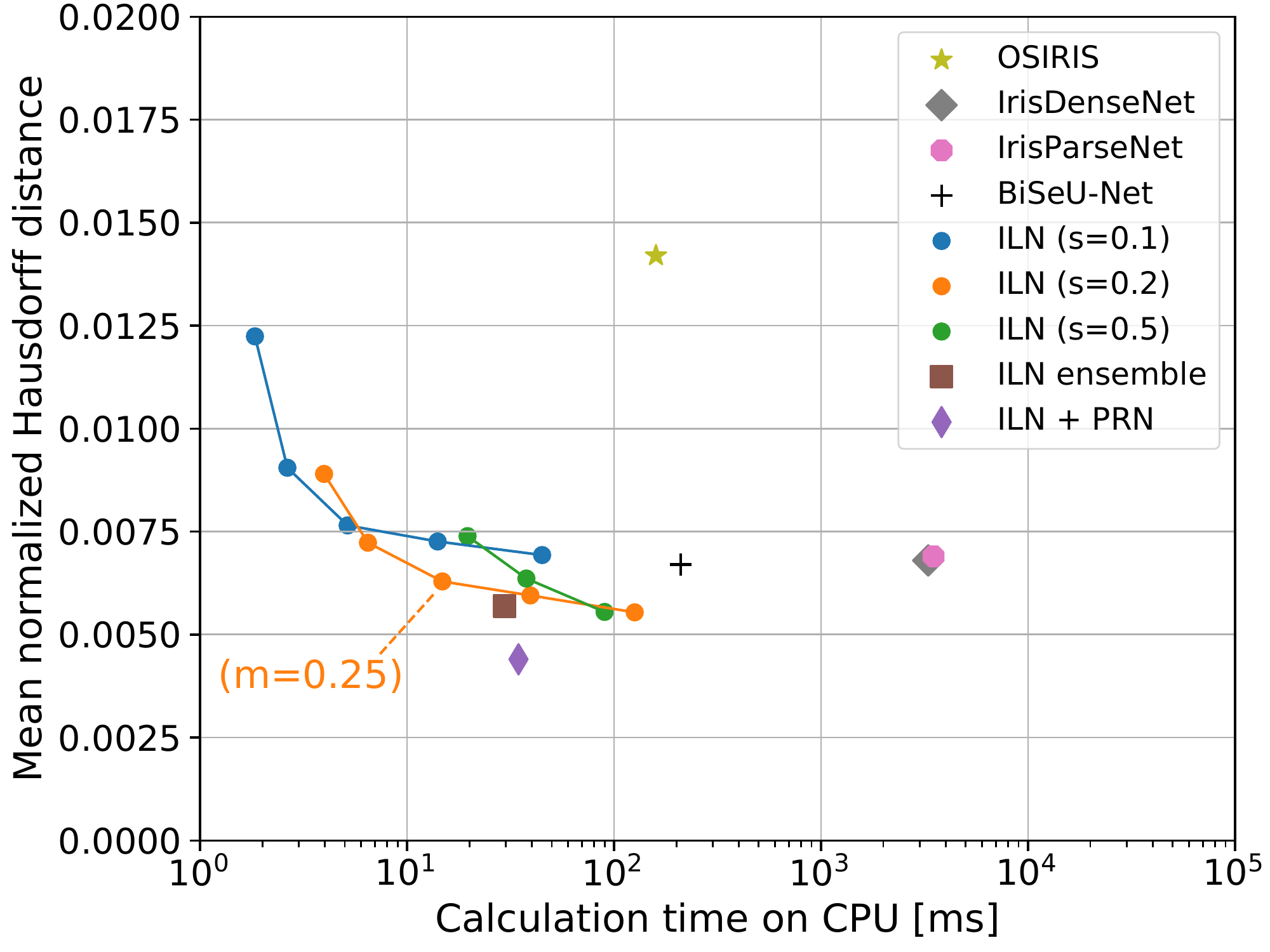}
\end{center}
   \caption{Results of ablation study for pupil localization using CASIA v4-thousand dataset. All points of proposed method (ILN) were faster than conventional ones while maintaining localization accuracies.}
\label{fig:result2}
\end{figure}

Figure \ref{fig:result2} shows the results of the ablation study. The vertical axis is the mean normalized Hausdorff distance of the pupil circle, and the horizontal axis is the computational time for one image on the CPU. For comparison, we plotted the localization results of OSIRIS v4.1 \cite{othman2016osiris} (yellow star), IrisParseNet with attention \cite{wang2020towards} (pink circle), IrisDenseNet \cite{arsalan2018irisdensenet} (black plus sign), and BiSeU-Net \cite{zhao2021detection} (gray diamond) for the same dataset. The blue, orange, and green lines are the results of ILN with the $s$ parameter set to (0.1, 0.2, 0.5), respectively. The case of $s$ = 0.2 had the best performance in terms of both speed and accuracy. The ILN model with ($s$, $m$) = (0.2, 0.25) achieved performance comparable with other CNN-based methods with a calculation time of 14.8 ms. The ensemble model showed a further accuracy improvement (brown square). The proposed PRN (purple diamond) achieved the best performance in terms of the pupil localization accuracy.

\begin{figure}[t]
\begin{center}
   \includegraphics[width=0.89\linewidth]{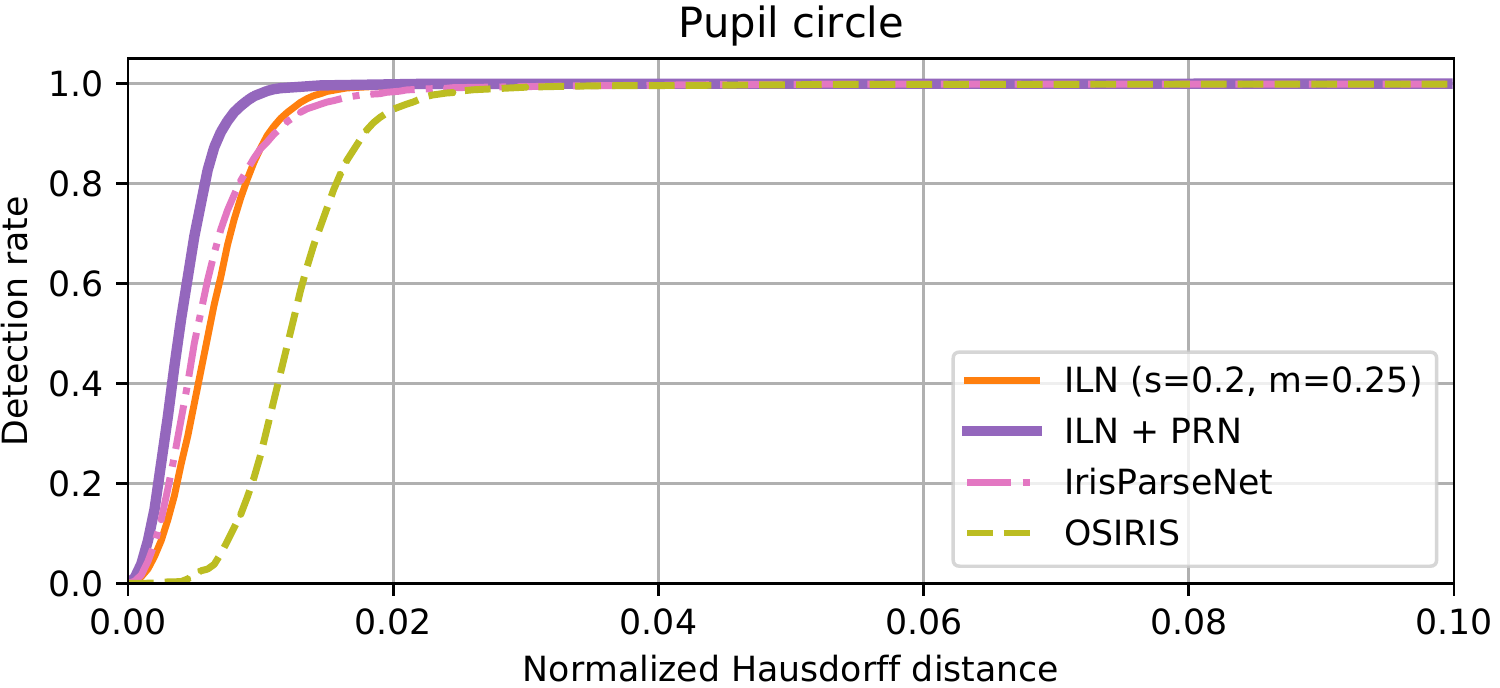}
   \includegraphics[width=0.89\linewidth]{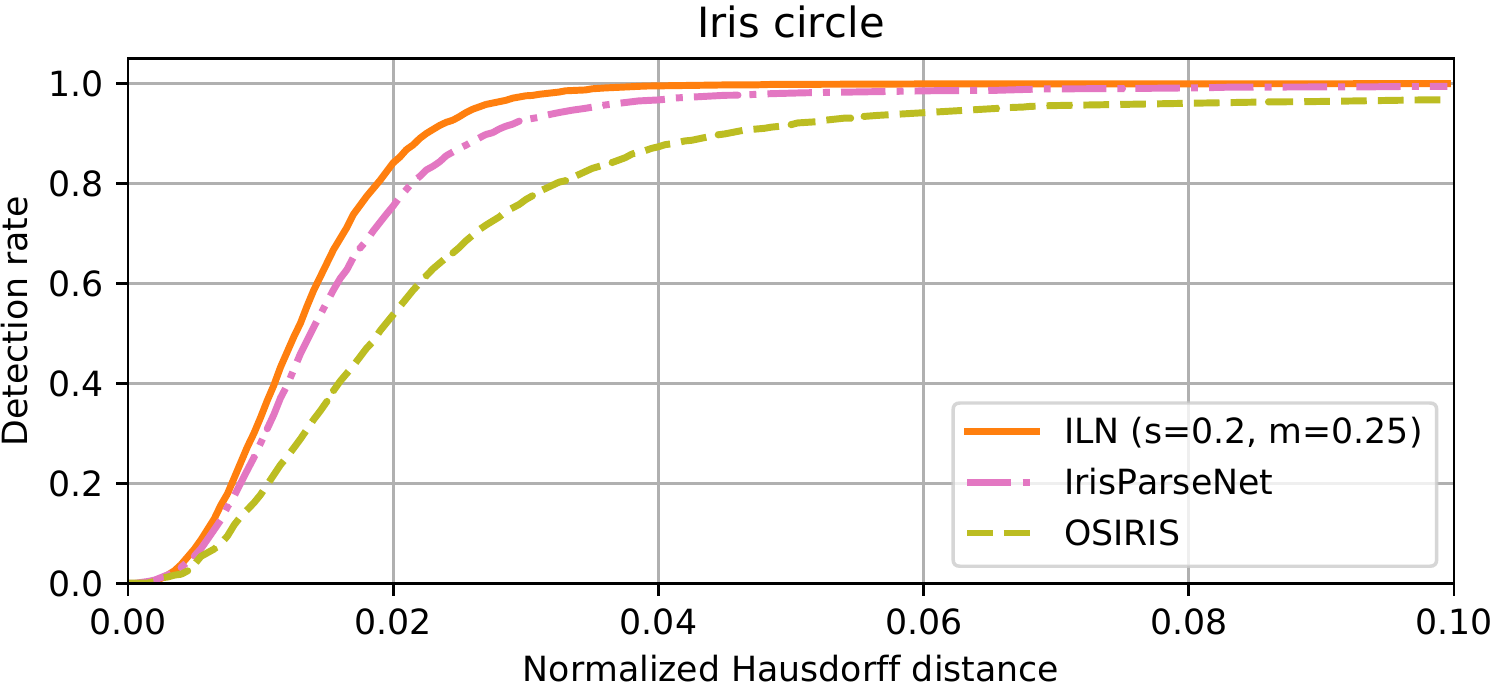}
\end{center}
   \caption{CED curves of pupil (upper) and iris (lower) circle localization for CASIA v4-thousand dataset.}
\label{fig:result-ciris}
\end{figure}

Figure \ref{fig:result-ciris} shows the cumulative error distribution (CED) curves of the detection rate for the pupil and iris circles. CED curves are the result of using the CASIA v4-thousand dataset. The horizontal axis is the normalized Hausdorff distance and the vertical axis is the cumulative detection rate. The closer the horizontal curve is to 0, the better the detection accuracy, and the faster the vertical axis converges to 1.0, the higher the robustness. We selected IrisParseNet and OSIRIS as the learning and non-learning based conventional methods. IrisParseNet had stable performance on both pupil and iris localization. In pupil detection, the performance of ILN was better than OSIRIS localization, and it was comparable 
to IrisParseNet. IrisParseNet had better performance than ILN at low detection rates, and ILN performed better at high detection rates. This means that the proposed method had better robustness than the conventional ones. The pupil circle refinement (ILN + PRN) had the best performance among all methods in terms of both detection accuracy and robustness. Regarding iris detection, the performance of the proposed method was comparable to that of IrisParseNet for the CED curves. Note that the iris localization result for the outer iris circle with ILN + PRN was the same as ILN because PRN re-localizes the pupil circle only. The outer iris circle is not refined by PRN.

To evaluate the localization accuracy for all output points, we investigated the distance between the localization results and the ground truth for each point. For the pupil and the iris, we used the center coordinate for comparison with other eyelid points. We calculated the Euclidean (L2) distance for all points, and we extracted the mean for each point. We used the test dataset of CASIA v4-thousand including images and their ground truth for this evaluation. 

\begin{figure}[t]
\begin{center}
   \includegraphics[width=0.89\linewidth]{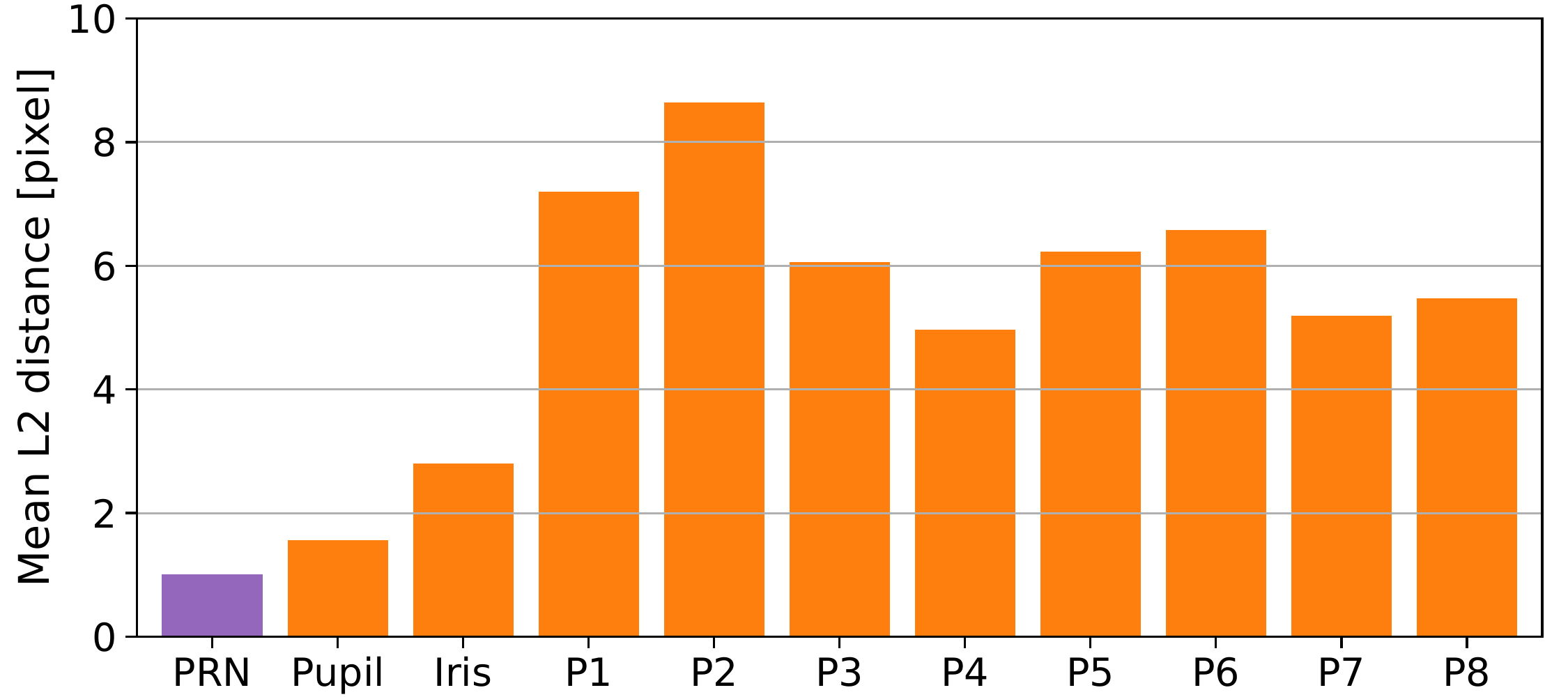}
\end{center}
   \caption{L2-distance errors of circle centers and eyelid points for CASIA v4-thousand dataset. PRN represents error of refined pupil center. Other errors are results of ILN.}
\label{fig:result-errors}
\end{figure}

\begin{table*}[t]
\caption{Dataset domain specific evaluation results for pupil and iris circle localizations. Each test dataset has same domain as corresponding training dataset. Methods were compared by processing time and mean normalized Hausdorff distance.}
\label{tab:dataset_result_p}
\begin{center}
\scalebox{0.88}{
\begin{tabular}{l|rr|cc|cc|cc}
\hline 
 & \multicolumn{2}{c|}{Time [ms]} & \multicolumn{2}{c|}{CASIA thousand} &  \multicolumn{2}{c|}{CASIA distance} & \multicolumn{2}{c}{IITD} \\
Methods & CPU & GPU & Pupil & Iris & Pupil & Iris & Pupil & Iris \\
\hline 
OSIRIS \cite{othman2016osiris} & 159.4 & -- & 0.0142 & 0.0272 & 0.0285 & 0.0821 & 0.0067 & 0.0221\\
CC-Net \cite{fang2020open}     & 48.9 & 4.4 & 0.1546 & 0.1979 & 0.1012 & 0.0926 & 0.0410 & 0.0385 \\ 
IrisParseNet \cite{wang2020towards} & 3494.8 & 15.7 & 0.0069 & 0.0171 & 0.0118 & 0.0234 & 0.0052 & 0.0149 \\
IrisDenseNet \cite{arsalan2018irisdensenet} & 3295.8 & 11.9 & 0.0068 & 0.0422 & 0.0061 & 0.0267 & \textbf{0.0041} & 0.0233 \\
BiSeU-Net \cite{zhao2021detection}  & 210.0  & 6.9 & 0.0067 & 0.0497 & 0.0139 & 0.0423 & 0.0045 & 0.0309\\
ILN (ours) & \textbf{14.8} & \textbf{3.2} & 0.0063 & \textbf{0.0132} & 0.0082 & \textbf{0.0184} & 0.0049 & \textbf{0.0129}\\
ILN+PRN (ours) & 34.5 & 5.6 & \textbf{0.0044} & -- & \textbf{0.0053} & -- & 0.0043 & -- \\
\hline 
\end{tabular}
}
\end{center}
\end{table*}

\begin{table*}[t]
\caption{Generalization performance for pupil and iris localizations. CNN-based methods (IrisParseNet, IrisDenseNet, BiSeU-Net, ILN and ILN-PRN) were trained by CASIA-thousand dataset only, and other five test datasets are not included in training. For only CC-Net, we used public pre-trained model. Methods were compared by mean normalized Hausdorff distance for pupil and iris circles.}
\label{tab:dataset_result_gene}
\begin{center}
\scalebox{0.88}{
\begin{tabular}{l|cc|cc|cc|cc|cc|cc}
\hline 
 & \multicolumn{2}{c|}{Training \& test} & \multicolumn{10}{c}{Test only} \\
\hline
 & \multicolumn{2}{c|}{CASIA-thousand} & \multicolumn{2}{c|}{CASIA-distance} & \multicolumn{2}{c|}{IITD} & \multicolumn{2}{c|}{MMU1} & \multicolumn{2}{c|}{MMU2} & \multicolumn{2}{c}{CASIA-twins}\\
Methods & Pupil & Iris & Pupil & Iris & Pupil & Iris & Pupil & Iris & Pupil & Iris & Pupil & Iris \\
\hline 
OSIRIS \cite{othman2016osiris} 
& 0.0142 & 0.0272 
& 0.0285 & 0.0821 
& 0.0067 & 0.0221 
& 0.0113 & 0.0168 
& 0.0138 & 0.0271 
& 0.0110 & 0.0203 \\
CC-Net \cite{fang2020open} 
& 0.1546 & 0.1979 
& 0.1012 & 0.0926 
& 0.0410 & 0.0385 
& 0.0228 & 0.0201 
& 0.0654 & 0.0634 
& 0.0515 & 0.0567 \\ 
IrisParseNet \cite{wang2020towards}
& 0.0069 & 0.0171
& 0.0705 & 0.0471
& 0.0156 & 0.1365
& 0.0400 & 0.0256
& 0.1126 & 0.0657
& 0.0467 & 0.0164 \\
IrisDenseNet \cite{arsalan2018irisdensenet}
& 0.0068 & 0.0422
& 0.0464 & 0.0377
& 0.0135 & 0.0557
& 0.0158 & 0.0168
& 0.0286 & 0.0551
& 0.0351 & 0.0272 \\
BiSeU-Net \cite{zhao2021detection}
& 0.0067 & 0.0497
& 0.0383 & 0.0721
& 0.0051 & 0.0444
& 0.0118 & 0.0310
& 0.0208 & 0.0628
& 0.0232 & 0.0453 \\
ILN (ours)
& 0.0063 & \textbf{0.0132}
& 0.0189 & \textbf{0.0364}
& 0.0077 & \textbf{0.0142}
& 0.0093 & \textbf{0.0156}
& 0.0113 & \textbf{0.0244}
& 0.0118 & \textbf{0.0133} \\
ILN+PRN (ours)
& \textbf{0.0044} & --
& \textbf{0.0107} & --
& \textbf{0.0050} & --
& \textbf{0.0064} & --
& \textbf{0.0069} & --
& \textbf{0.0085} & --\\
\hline 
\end{tabular}
}
\end{center}
\end{table*}

Figure \ref{fig:result-errors} shows errors for all detected circle centers and eyelid points. The eyelid points (from $P_1$ to $P_8$) had different error values from each other due to the difference in the annotation accuracy. In particular, the eye corner points ($P_1$ and $P_2$) tended to be annotated at different positions by different annotators because these points were difficult to determine for each iris image. On the other hand, pupil and iris errors were smaller than those of the other eyelid points because of the weight of the loss function. The pupil error for PRN was further improved by the refinement. Although the localization accuracy of all circle centers and points can be controlled by the weight of the loss function or the refinement, we focused on the pupil circle accuracy because it strongly affects the iris recognition performance.

\subsection{Dataset dependency}

We evaluated the dataset dependency of the proposed method using three of the above datasets: CASIA v4-thousand, CASIA v4-distance, and IITD \cite{casia, kumar2010comparison}. For each dataset, each model was trained and evaluated on the basis of the splits in Table \ref{tab:dataset}. The evaluation index was the mean normalized Hausdorff distance of the pupil and iris circles. The same as in the ablation study, OSIRIS v4.1 \cite{othman2016osiris}, IrisParseNet with attention \cite{wang2020towards}, IrisDenseNet \cite{arsalan2018irisdensenet} and BiSeU-Net \cite{zhao2021detection} were used as the conventional methods for comparison. To train these conventional networks, we used the same data augmentation as our method, and the number of iteration was set to 50,000. For IrisDenseNet and BiSeU-Net, we set iris and pupil regions as output channels. The other parameters were the same as those used by Wang \etal \cite{wang2020towards}. The ground truths of the segmentation masks were created by using our annotated circles and points. We fixed the parameters of the proposed ILN at ($s$, $m$) = (0.2, 0.25).

Table \ref{tab:dataset_result_p} shows the pupil and iris localization results for dataset dependency. Each value is the mean Hausdorff distance between the localized circle and ground truth circle. Calculation time is measured for a single image using Intel Xeon CPU E3-1280 v5 (CPU) or NVIDIA GeForce RTX 3080 (GPU). For pupil localization, the proposed method with pupil refinement (ILN + PRN) had the best performance for two of the datasets and worked within a computational time of 34.5 ms on CPU, which was faster than the conventional methods. For IITD, ILN + PRN was second in terms of performance. These results confirm that the proposed method had better performance than the other methods with the fastest computational time for all three datasets. Note that the pupil error of the IrisParseNet result for CASIA v4-distance was larger than the 0.0069 reported by Wang \etal \cite{wang2020towards} due to the different dataset splits and annotation. The iris localization performance of the proposed ILN outperformed the other methods for all three datasets. Note that ILN + PRN is not included in the iris results because PRN refines only pupil circle. Thus, the iris localization result of ILN + PRN was the same as that of ILN. 

\subsection{Generalization performance}

\begin{figure}[t]
\begin{center}
   \includegraphics[width=0.99\linewidth]{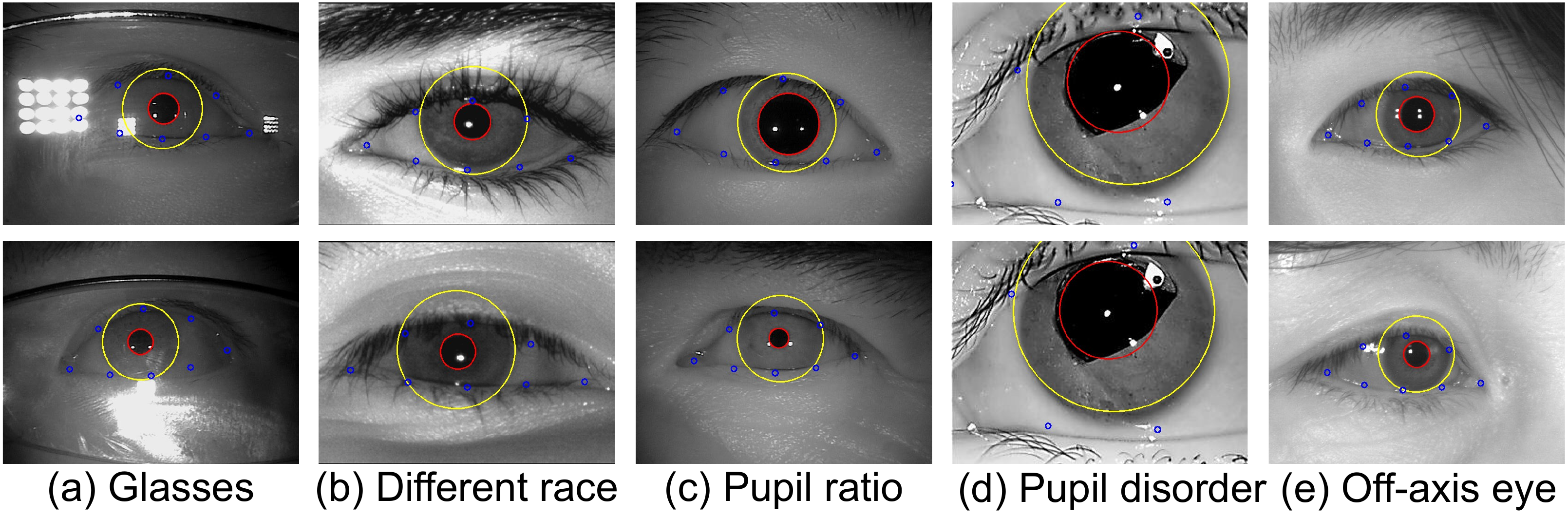}
\end{center}
   \caption{Results for complex iris images. Proposed method robustly localizes pupil and iris circles for each condition.}
\label{fig:loc}
\end{figure}

The performance of CNN-based methods tends to depend on the domain of the training data, and their generalization performance to other domains than the training domain is often not guaranteed. Therefore, to investigate the generalization performance of the proposed method, we evaluated the performance of models trained only on CASIA v4-thousand using other datasets. We used CASIA v4-distance, IITD, MMU1, MMU2 and CASIA v4-twins as evaluation datasets. OSIRIS and CC-Net were applied directly to the publicly available models, while IrisParseNet, IrisDenseNet, BiSeU-Net and the proposed method were applied to models trained using CASIA-thousand. 

Table \ref{tab:dataset_result_gene} shows the localization results of the generalizability evaluation. The methods were compared on the basis of the mean normalized Hausdorff distance. We confirmed that the proposed method had better generalization performance than the segmentation methods for all datasets. The non-training method, OSIRIS, had a higher generalization performance, while the learning-based method caused the generalization performance to degrade in domains that differed from the training data. The reason that the proposed landmark detection was less affected by the domain is considered to be that it uses only edges and avoids using light-sensitive areas compared with the segmentation methods.

Figure \ref{fig:loc} shows the detection results for complex conditions: glasses, different race from training dataset, different pupil ratio, pupil disorder and off-axis eye for the case of using a generalized model. The proposed method worked robustly for glasses, different race and different pupil ratio. For pupil disorder, it localized circle-like shapes as pupil, and the difference in the results between the two images was not large. For off-axis eye, the proposed method approximated ellipse eyes by circles. These results confirm that our model stably localizes iris circles for complex images.

\subsection{Extension to ellipse localization}

\begin{figure}[t]
\begin{center}
   \includegraphics[width=0.99\linewidth]{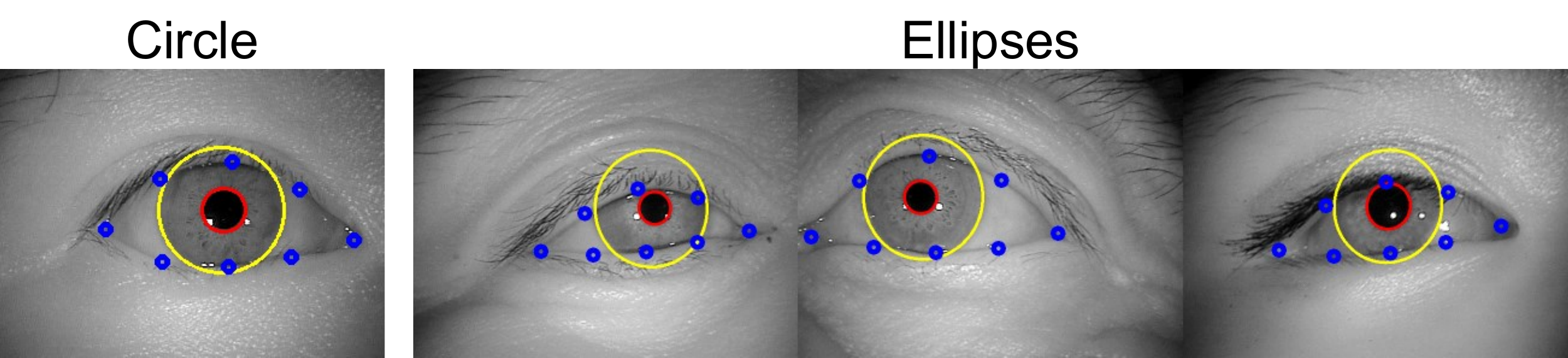}
\end{center}
   \caption{Results of ellipse localization. Proposed method can be extended to ellipse fitting with the additional of target parameters.}
\label{fig:ellipse}
\end{figure}

Our proposed method can easily be extended to ellipse localization. For circles, we define parameters as ($x, y, r$). In the case of ellipses, parameters are extended to ($x, y, a, b, \theta$), where ($x, y$) is the center coordinate, ($a, b$) is the length of the main and sub axes, and $\theta$ is the ellipse rotation angle. Figure \ref{fig:ellipse} shows ellipse localization results. Since we had no ground truth for the ellipses, we trained an ellipse localization model using ellipses generated by data augmentation with affine transformation for annotated circles. As shown in Figure \ref{fig:ellipse}, the ellipse localization worked correctly for ellipses. However, we had no evaluation data for ellipses, so ground truth annotation for ellipses and numerical evaluation are needed as future work.

\subsection{Iris recognition}
\label{sec:recognition}

We evaluated the performance of iris recognition with the proposed iris localization method. To reduce the recognition performance dependency on the dataset domain, we used no CNN-based recognition methods but OSIRIS v4.1 \cite{othman2016osiris} as the recognition engine. We split the processing of OSIRIS into localization, segmentation, and recognition. We replaced the localization and segmentation processing of OSIRIS with the proposed method. For input images, we first localized the pupil and iris circles with the proposed method. After the localization, we made segmentation maps using the eyelid points and pixel values of the iris regions. Finally, the localization and segmentation results were used for the recognition processing of OSIRIS. CASIA v4-distance was used for training and evaluation. We evaluated the recognition performance using the detection error tradeoff (DET) curve for five localization settings: OSIRIS localization, IrisDenseNet, ILN, ILN + PRN with an eyelid mask, and ILN + PRN with all masks. The all masks included eyelid, eyelash, and specular regions. The reason of selecting IrisDenseNet is that it is the most accurate method for pupil localization in the conventional CNN-based methods.

\begin{figure}[t]
\begin{center}
   \includegraphics[width=0.99\linewidth]{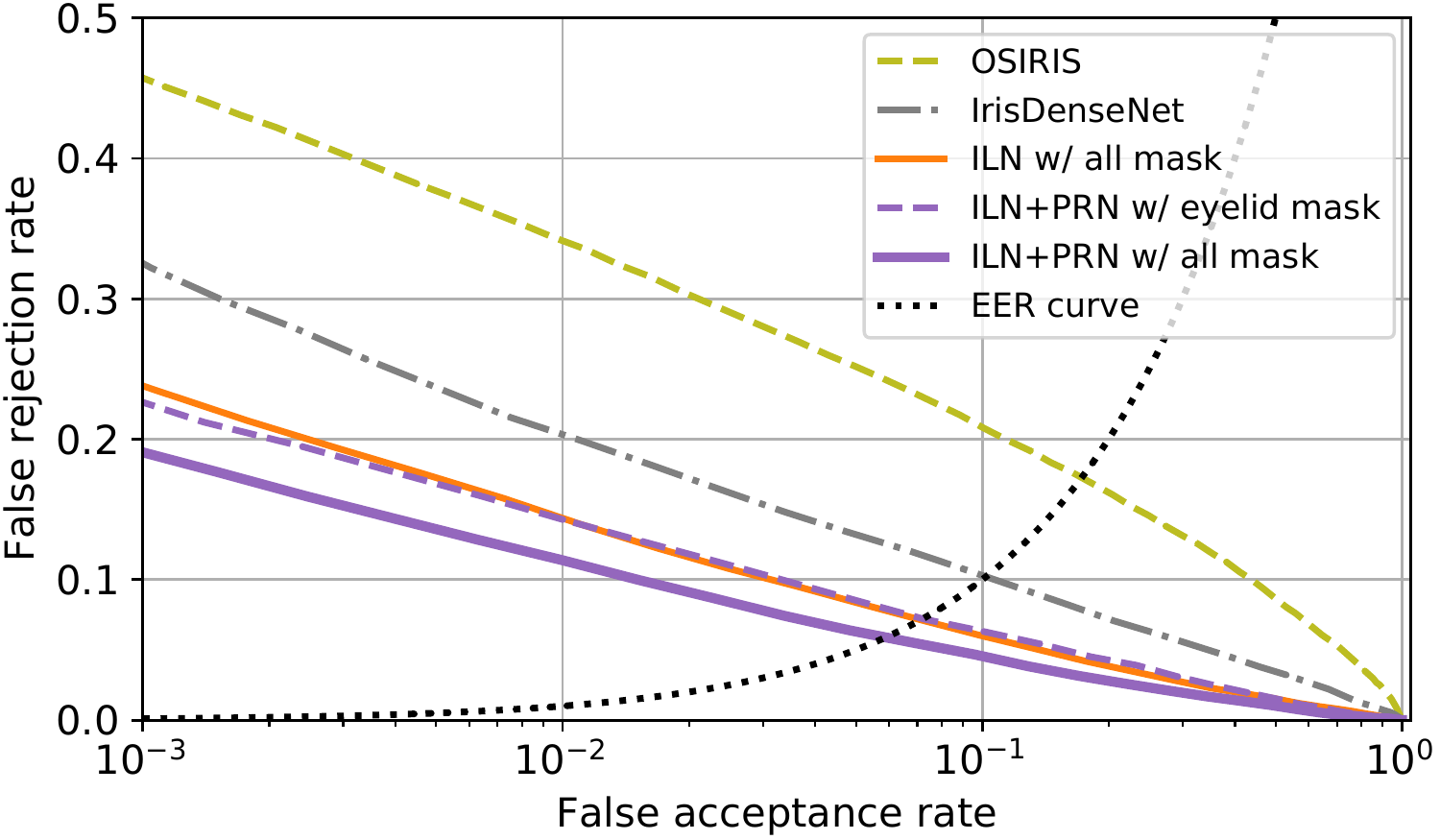}
\end{center}
   \caption{Detection error tradeoff curves for iris recognition. These curves were calculated using CASIA v4-distance dataset.} 
\label{fig:result-rec}
\end{figure}

Figure \ref{fig:result-rec} shows DET curves for iris recognition. The horizontal axis is the false acceptance rate, and the vertical axis is the false rejection rate. Since both axes express error rates, the closer the line is to the lower left, the better the performance. The black curve represents the equal error rate (EER). EERs were improved to 0.072 for ILN and to 0.062 for PRN, compared with 0.169 for OSIRIS and 0.098 for IrisDenseNet. EER improvement by the pupil refinement was comparable to that of the eyelash and specular region masks. These results confirmed that the localization accuracy was most important for recognition, and it improved the recognition performance significantly.


\section{Conclusion}
We have proposed a segmentation-free efficient iris localization method using an iris circle detector based on a deep regression network. To achieve more efficient iris localization than the conventional iris segmentation methods, the proposed ILN has directly localized pupil and iris circles with eyelid points from down-sampled iris images. We have also introduced PRN to improve the pupil localization accuracy. Experimental results have shown that the proposed method extracted iris regions in 34.5 ms on a CPU while achieving better localization performance than the conventional iris segmentation-based methods. We have also confirmed that our method improves iris recognition accuracies compared with the conventional method.


{\small
\bibliographystyle{ieee_fullname}
\bibliography{egbib}
}

\end{document}